%% file: acl_latex.tex
\pdfoutput=1

\documentclass[11pt]{article}

\usepackage[final]{acl}

\usepackage{times}
\usepackage{latexsym}

\usepackage[T1]{fontenc}

\usepackage[utf8]{inputenc}

\usepackage{microtype}

\usepackage{inconsolata}
\usepackage{amssymb}
\usepackage{mathtools}
\usepackage{amsthm}
\usepackage{amsfonts} 
\usepackage{graphicx}
\usepackage{amsmath,bm} 
\usepackage{hyperref}
\usepackage{url}
\usepackage{booktabs}
\usepackage{multirow}
\usepackage{colortbl}
\usepackage[export]{adjustbox}
\usepackage{wrapfig}
\usepackage{pifont}
\newcommand{\cmark}{\ding{51}}%
\newcommand{\xmark}{\ding{55}}%
\title{DLP-LoRA: Efficient Task-Specific LoRA Fusion with a Dynamic, Lightweight Plugin for Large Language Models}

\author{Yuxuan Zhang$^{1,2}$ \quad Ruizhe Li$^{1}\thanks{Corresponding author: \texttt{ruizhe.li@abdn.ac.uk}.}$ 
\\
$^1$Department of Computing Science, University of Aberdeen\\ $^2$Aberdeen Institute of Data Science and Artificial Intelligence, South China Normal University
}

\begin{document}
\maketitle
\begin{abstract}
Recent advancements in Large Language Models (LLMs) have achieved robust performance across diverse tasks, but fine-tuning these models for specific domains remains resource-intensive. Parameter-Efficient Fine-Tuning (PEFT) methods like Low-Rank Adaptation (LoRA) address this challenge by fine-tuning a small subset of parameters. However, existing methods for fusing multiple LoRAs lack dynamic fusion based on contextual inputs and often increase inference time due to token-level operations. We propose DLP-LoRA, a Dynamic Lightweight Plugin that employs a mini-MLP module with only 5M parameters to dynamically fuse multiple LoRAs at the sentence level rather than the token level using top-$p$ sampling strategies for possible LoRAs. This approach reduces inference time to less than 2x that of a single LoRA inference by leveraging parallel computation. Evaluations across 26 tasks, including multiple-choice questions and question answering, demonstrate that DLP-LoRA achieves an average accuracy of 91.9\% on multiple-choice datasets and significant improvements in BLEU, ROUGE-1 and ROUGE-L scores (54.1\%, 43.5\% and 40.8\%) on QA datasets, outperforming many LoRA baselines under different LLMs backbones. DLP-LoRA effectively balances performance and efficiency, making it a practical solution for dynamic multi-task adaptation in LLMs.
\end{abstract}

\input{sections/introduction}

\input{sections/background}

\input{sections/methodology}

\input{sections/experiment}

\input{sections/discussion}

\input{sections/related_work}

\section{Conclusion}
We introduced DLP-LoRA, a dynamic and lightweight plugin that employs a mini-MLP module with only 5 million parameters to dynamically fuse multiple LoRAs at the sentence level using top-$p$ sampling strategies. Our comprehensive evaluation across 17 MCQ tasks and 9 QA tasks demonstrates that DLP-LoRA not only closely matches the performance of individually fine-tuned single LoRAs but also surpasses them on certain tasks, all while incurring less than twice the inference time. Through detailed discussions and ablation studies, we have shown that DLP-LoRA effectively balances performance and efficiency in multi-task learning, making it a practical solution for dynamic multi-task adaptation in LLMs.

\section*{Limitations}

Our evaluation of DLP-LoRA was primarily conducted on LLM backbones ranging from 1.5 billion to 8 billion parameters, constrained by the computational limitations of our GPU resources. Consequently, we were unable to assess the performance of DLP-LoRA on larger models such as Qwen-2.5 32B~\citep{hui2024qwen2} and LLaMA-3.1 70B~\citep{dubey2024llama}, which may exhibit different behaviors and performance characteristics. Additionally, when composite tasks include a higher proportion of MCQ datasets, DLP-LoRA tends to assign higher probabilities to the specific MCQ LoRA, potentially limiting its ability to effectively fuse and utilize QA LoRAs. This tendency might restrict the diversity of generated outputs and the fusion capabilities of DLP-LoRA across a broader range of tasks.

\bibliography{custom}

\appendix

\input{sections/appendix}

\end{document}

%% file: sections/introduction.tex
\section{Introduction}

Recent advancements in Large Language Models (LLMs) such as LLaMA 3.1~\citep{dubey2024llama}, Qwen 2.5~\citep{qwen2.5}, and Gemma 2~\citep{team2024gemma} have led to robust and superior performance across multiple benchmarks~\citep{muennighoff2022mteb,llm-perf-leaderboard,open-llm-leaderboard-v2}. These models have demonstrated remarkable capabilities in diverse areas, including code generation~\citep{qwen}, mathematical reasoning~\citep{ahn2024large}, and question answering~\citep{achiam2023gpt}. Despite these achievements, fine-tuning all parameters of such large models for specific domains remains resource-intensive and time-consuming.

Parameter-Efficient Fine-Tuning (PEFT) methods~\citep{houlsby2019parameter,xu2023parameter} address this challenge by enabling the fine-tuning of a small subset of parameters, thereby improving performance in various applications like multi-task learning~\citep{xu2024meteora,kong2024lora}, multilingual summarisation, and transfer learning~\citep{whitehouse-etal-2024-low,zhao2024adamergex}. One prominent PEFT approach is Low-Rank Adaptation (LoRA)~\citep{hu2021lora}, which fine-tunes low-rank matrices to capture domain-specific knowledge and merges them with pre-trained LLMs.

To enhance the multi-task learning capabilities of LLMs, several methods have been proposed to fuse task-specific LoRAs, including Arrow~\citep{ostapenko2024towards}, LoRAHub~\citep{huang2024lorahub} and MeteoRA~\citep{xu2025meteora}. These approaches primarily use learnable gating networks or multiple iterations to adapt and combine multiple LoRAs. For instance, MeteoRA~\citep{xu2024meteora} introduces 7 token-level gating networks to all attention and MLP layers for dynamic LoRA fusion.

However, most of these methods lack the ability to dynamically fuse LoRAs based on contextual prompt inputs during inference. They either require manual selection before combining LoRAs or necessitate additional fine-tuning of embedded gating networks when new tasks are introduced. Moreover, existing LoRA mixture strategies like MeteoRA focus on token-level Mixture-of-Experts (MoE) gating across all attention heads and MLP layers, which significantly increases inference time for next-token generation. Observations from prior studies~\citep{xu2025meteora,lin2024teamlora,muqeeth2024learning} indicate that within the same sentence of a task, the same LoRA is consistently assigned to each token. This suggests that token-level LoRA MoE might be unnecessary and computationally inefficient.

In this paper, we propose a Dynamic Lightweight Plugin for LoRA fusion (DLP-LoRA), which employs a lightweight mini-MLP module to dynamically fuse multiple LoRAs based on top-$p$ sampling strategies on the sentence level. This mini-MLP plugin, containing only 5M parameters, is fast to train for multi-task classification and easily adaptable to new domains, such as increasing task numbers from 50 to 100. By leveraging sentence-level LoRA selection and fusion guided by the mini-MLP plugin, DLP-LoRA requires less than 2x the inference time compared to manually selecting and loading a single LoRA and different LoRA baselines equipped with dynamic fusion methods, making it comparable in efficiency.

We evaluate DLP-LoRA across 26 tasks, including 18 multiple-choice question (MCQ) datasets spanning mathematical QA, logical reasoning, language identification, and reading comprehension, as well as 8 question-answering (QA) datasets focused on summarisation, machine translation, and open-domain QA. Under comparable inference times to single LoRA setups and different dynamic LoRA baselines, DLP-LoRA achieves an average accuracy of \textbf{91.9\%} across the 18 MCQ datasets and average BLEU, ROUGE-1, and ROUGE-L scores of \textbf{54.1}, \textbf{43.5}, and \textbf{40.8}, respectively, across the 8 QA datasets. These evaluations are conducted using Qwen-2 1.5B, Qwen-2 7B, LLaMA-2 7B, and LLaMA-3 8B backbones. Additionally, our model demonstrates relative improvements of \textbf{92.95\%} and \textbf{13.2\%} for the MCQ and QA tasks, respectively, compared to different LLM backbones under composite task settings. With DLP-LoRA, the inference speed of the LLaMA-2 7B backbone is improved by average \textbf{353.8\%} compared to different dynamic LoRA baselines. Our case studies further illustrate that sentence-level DLP-LoRA effectively balances the trade-off between multi-LoRA inference and fusion.

In summary, our contributions are threefold:
\begin{itemize}
    \item We introduce DLP-LoRA, a dynamic and lightweight plugin for multi-LoRA selection and fusion that is fast to train and easily adaptable to new domains.
    \item By employing sentence-level multi-LoRA selection and fusion, DLP-LoRA leverages parallel CUDA acceleration, achieving less than 2x the inference time compared to single LoRA inference and outperforming token-level MoE gating routers in efficiency.
    \item Through extensive evaluations on 26 tasks including MCQ and QA, DLP-LoRA significantly improves accuracy, BLEU, ROUGE-1 and ROUGE-L compared to different SOTA LoRA baselines under single and composite task settings.
\end{itemize}

%% file: sections/background.tex
\section{Background}
\begin{figure*}[tb]
    \centering
    \includegraphics[width=\textwidth]{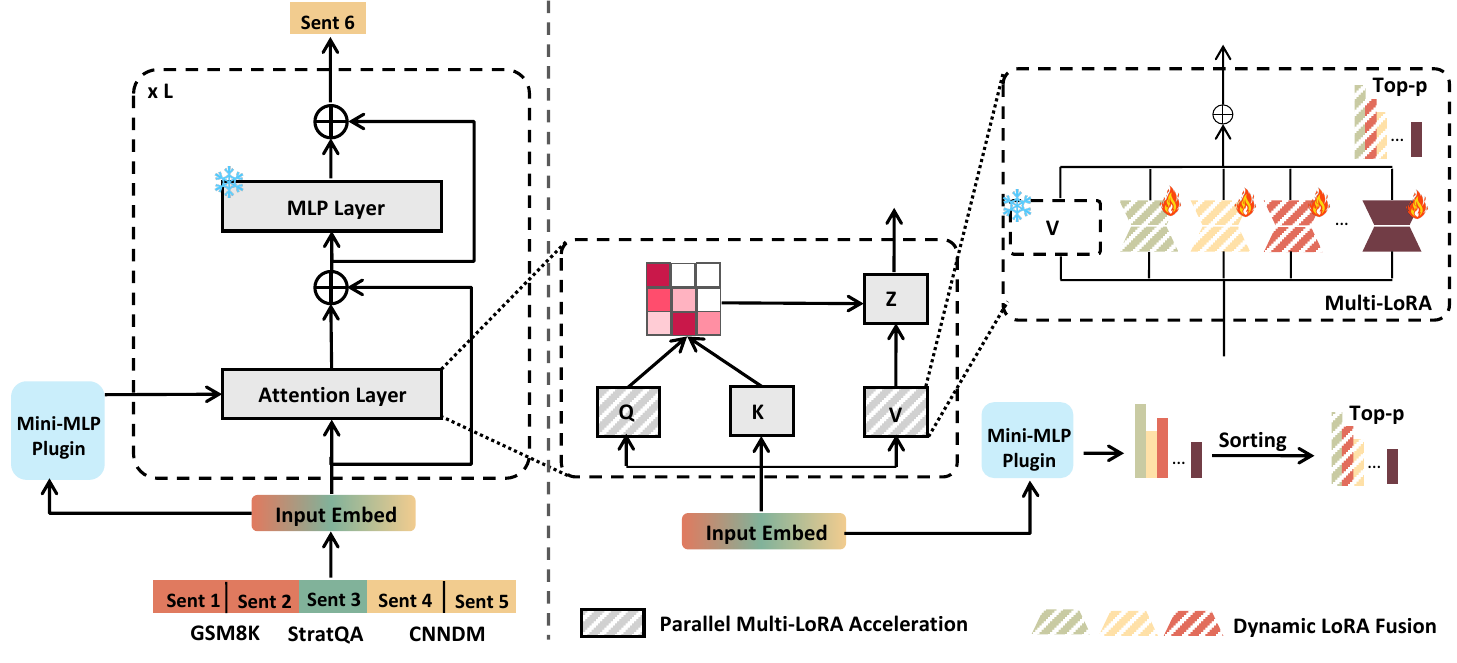}
    \caption{DLP-LoRA framework: different LoRAs will be activated based on the input task and sentence via mini-MLP plugin. When Top-$p$ sampling is used via the mini-MLP plugin, multiple LoRAs will be sampled and fused with probability $p$ as the threshold. DLP-LoRA fusion is only enabled once the first token of every new sentence is generated.}
    \label{fig:DLP_LoRA}
    \vspace{-1em}
\end{figure*}
\paragraph{Low-Rank Adaptation (LoRA).}
LoRA~\citep{hu2021lora} fine-tunes LLMs efficiently by freesing most pre-trained weights and adding low-rank matrices to specific layers, notably within Transformer attention projections (and recently, MLP layers~\citep{dou2024loramoe,li2024mixlora}). Given a weight matrix \( \boldsymbol{W} \in \mathbb{R}^{h \times d} \), LoRA introduces matrices \( \boldsymbol{A} \in \mathbb{R}^{h \times r} \) and \( \boldsymbol{B} \in \mathbb{R}^{r \times d} \) with \( r \ll \min(h,d) \), modifying the weight as:
\begin{equation}
   \boldsymbol{W}' = \boldsymbol{W} + \boldsymbol{AB}. 
\end{equation}
For an input \( \boldsymbol{x} \), the output becomes \( \boldsymbol{h} = \boldsymbol{xW} + \boldsymbol{xAB} \). This approach leverages the insight that fine-tuning updates often lie in a low-dimensional subspace, drastically reducing trainable parameters (sometimes by up to 10,000×) while keeping inference efficient, since the low-rank matrices can be merged with the original weights after training.

\paragraph{Multi-task LoRA Mixture.} 
A single LoRA adapter is tailored to one downstream task, limiting its utility to that particular application. To enable multi-task handling, one approach fine-tunes a single adapter on a combined dataset, but this can dilute domain-specific knowledge~\citep{lin2024teamlora}. Alternatively, individual LoRA adapters can be treated as modular components. Some architectures combine multiple adapters via a learnable weighted sum~\citep{huang2023lorahub} or unified CUDA memory pools~\citep{sheng2023s}, though these often require manual selection and additional few-shot or in-context learning. A more dynamic method, as seen in MeteoRA~\citep{xu2025meteora}, uses a token-level Mixture-of-Experts framework with a trainable gating mechanism across layers to automatically fuse different LoRAs. However, the inclusion of a trainable gating module at every attention and MLP layer with token-level routing significantly increases inference time compared to single LoRA inference. This performance drawback remains substantial even with the development of GPU kernel acceleration methods.

%% file: sections/methodology.tex
\section{Methodology}

Our proposed DLP-LoRA framework comprises three key components: a lightweight mini-MLP plugin $\mathcal{C}_{\text{MLP}}$, a base LLM backbone $\mathcal{M}$, and a set of $N$ fine-tuned LoRA modules $L_{\{1...N\}}$ corresponding to different tasks $\mathcal{D}_{\{1...N\}}$, as illustrated in Figure~\ref{fig:DLP_LoRA}. Initially, we train the mini-MLP classifier $\mathcal{C}_{\text{MLP}}$ on these tasks to achieve high task classification accuracy (we evaluate 26 tasks in this work; see Appendix~\ref{app:26_datasets} for details). Once trained, the LLM backbone $\mathcal{M}$ utilises the mini-MLP plugin to dynamically fuse the appropriate fine-tuned LoRAs $L_{\{1...N\}}$ at the sentence level, enabling efficient multi-task learning.

\subsection{Lightweight Multi-task Classification Plugin}

Previous methods that perform token-level task classification and routing within the LLM backbone, by injecting a trainable gating network at each attention and MLP layer, are computationally intensive and inefficient during inference~\citep{xu2025meteora}. Observing that most tokens within a sentence typically pertain to the same task~\citep{xu2025meteora,lin2024teamlora,muqeeth2024learning}, we propose a more efficient sentence-level task classification approach. Specifically, we introduce an off-the-shelf 4-layer mini-MLP plugin $\mathcal{C}_{\text{MLP}}$ that requires training only once on the sentence level for the selected $N$ tasks.

Given $N$ distinct tasks $\mathcal{D}_{\{1...N\}}$ and a collection of $M$ sentences $\mathcal{S}_{\{1...M\}} \in \mathcal{D}_n$, our lightweight 4-layer $\mathcal{C}_{\text{MLP}}$ encodes each input sentence $\mathcal{S}_m$ using a specific tokenizer (we utilise the ALBERT tokenizer~\citep{lan2019albert} in this work) and classifies $\mathcal{S}_m$ to the correct task $\mathcal{D}_n$:
\begin{equation}
    \mathcal{Y}_n = \mathcal{C}_{\text{MLP}}(\mathcal{S}_m), \quad \text{where} \quad \mathcal{Y}_n \in \mathcal{D}_{\{1...N\}},
\end{equation}

\subsection{Dynamic LoRA Fusion}

Once the $\mathcal{C}_{\text{MLP}}$ classifier is well-trained on the tasks $\mathcal{D}_{\{1...N\}}$, it serves as a plugin to the LLM backbone $\mathcal{M}$ for dynamically fusing multiple LoRAs $L_{\{1...N\}}$ at the sentence level. For the current input sentence $\mathcal{S}_m \in \mathcal{D}_n$, we consider the first token $\text{w}_1$ and the previous contextual history $\mathcal{H}_{\{1...k\}}$. We employ a top-$p$ sampling scheme via $\mathcal{C}_{\text{MLP}}$ to dynamically select the possible LoRAs to fuse, using probability $p$ as the threshold:
\begin{equation}\small
    \mathcal{I}_p = \{\mathcal{Y}_{\{1...R\}} \mid \text{w}_1 \in \mathcal{S}_m, \mathcal{H}_{\{1...k\}}\}, \quad \text{where} \quad \mathcal{Y}_r \geq p.
\end{equation}

Using the set $\mathcal{I}_p$ for the current sentence $\mathcal{S}_m$, we fuse the selected LoRAs based on normalised weights obtained via a softmax function:
\begin{equation}
    \mathcal{W}_m = \text{Softmax}(\mathcal{I}_p) = \{w_1, \ldots, w_R\}.
\end{equation}
Importantly, the $\mathcal{C}_{\text{MLP}}$ classifier is only activated when the first token $\text{w}_1$ of the current sentence $\mathcal{S}_m$ is generated, leveraging the contextual information $\mathcal{H}_{\{1...k\}}$. This approach significantly accelerates the inference time of $\mathcal{M}$ compared to token-level gating network classification~\citep{xu2025meteora}, as it avoids the overhead of per-token classification.

\subsection{Parallel Multi-LoRA Acceleration}

Beyond the efficiency gained from sentence-level LoRA sampling and fusion, which avoids the inefficiency of repetitive per-token LoRA classification, a significant advantage of our approach is the ability to fully exploit parallel multi-LoRA acceleration.

Given $N$ fine-tuned LoRAs, we construct two tensors $\boldsymbol{A} \in \mathbb{R}^{N \times h \times r}$ and $\boldsymbol{B} \in \mathbb{R}^{N \times r \times d}$, which are allocated contiguously in High Bandwidth Memory (HBM). In contrast to token-level LoRA classification and forward computation, where each token in the batch operates independently, limiting the effectiveness of General Matrix Multiplication (GEMM) optimisations in frameworks like PyTorch, our sentence-level LoRA classification removes the independence constraints among tokens within a sentence. By iterating over all $N$ LoRAs using a hash table stored in HBM, we retrieve the sampled LoRAs $\mathcal{I}_p$ based on top-$p$ sampling and their corresponding weights $\mathcal{W}_m$. Subsequently, all sampled LoRAs are fused into the original layer-wise weights $\boldsymbol{W}$ of the LLM as follows:
\begin{equation}\small
\begin{split}
    \underbrace{\left[\Delta \boldsymbol{o}_{1}, \ldots, \Delta \boldsymbol{o}_{BM}\right]}_{B \times M} &= \sum_{R} \boldsymbol{W}^{B \times M \times R} (( \underbrace{\left[ \boldsymbol{x}_{1}, \ldots, \boldsymbol{x}_{BMR} \right]}_{B \times M \times R} \times \\ & \underbrace{\left[ \boldsymbol{A}_{1}, \ldots, \boldsymbol{A}_{BMR} \right]}_{B \times M \times R} ) \times \underbrace{\left[ \boldsymbol{B}_{1}, \ldots, \boldsymbol{B}_{BMR} \right]}_{B \times M \times R} )
\end{split}
\end{equation}
where $B$ is the batch size, $M$ is the number of sentences, $R$ is the number of sampled LoRAs, and $\boldsymbol{x}$ represents the encoded representation of the first token of each input sentence $\mathcal{S}_m$. Normally, $M$ is significantly smaller than the token numbers during finetuning. Leveraging this parallel multi-LoRA acceleration, our DLP-LoRA achieves an inference time that is on average only 1.20x slower than single LoRA inference compared with 2.62x slower of MeteoRA (see Section~\ref{sec:exp_results} for detailed comparisons).

%% file: sections/experiment.tex
\section{Experiments}
\begin{figure*}[tb]
    \centering
    \includegraphics[width=\textwidth]{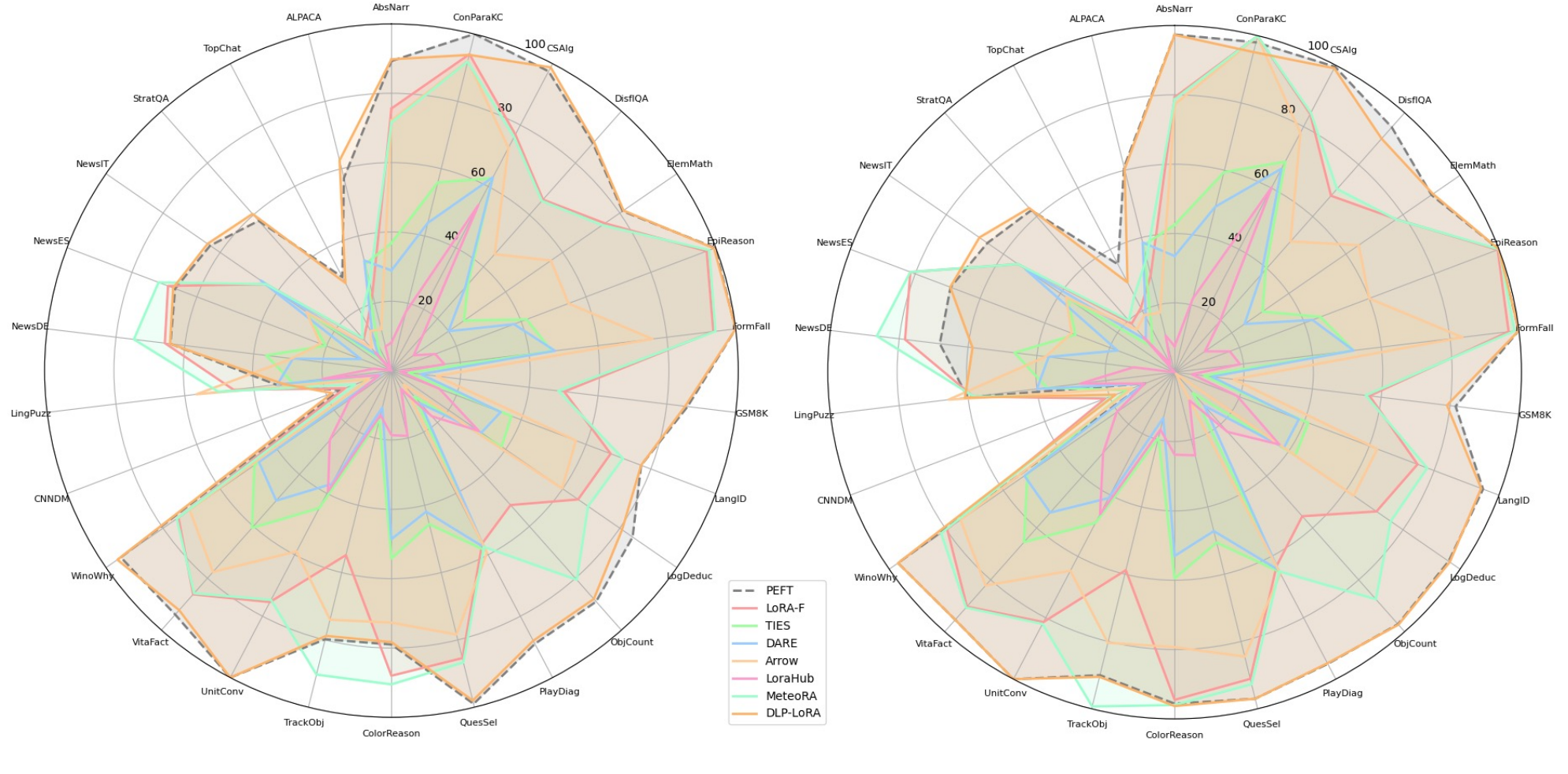}
    \caption{The performance of DLP-LoRA compared to 7 LoRA baselines using Qwen-2 1.5B (left) and LLaMA-3 8B (right) backbones across 26 tasks. See Appendix~\ref{app:all_results} for more results using Qwen-2 7B and LLaMA-2 7B LLMs backbones.}
    \label{fig:evaluation_26_llama2_7B}
\end{figure*}
\subsection{Experimental Setup}

\begin{table*}[tb]
    \centering
    \resizebox{\textwidth}{!}{
    \begin{tabular}{lccccc}
        \toprule
        \textbf{Models} &  \textbf{Accuracy}$\uparrow$ & \textbf{BLEU}$\uparrow$ & \textbf{ROUGE-1}$\uparrow$ & \textbf{ROUGE-L}$\uparrow$\\
        \midrule
        \textcolor{gray}{\textbf{PEFT (Ref.)}} & \textcolor{gray}{90.4 / 93.3 / 90.1 / 95.5} & \textcolor{gray}{51.0/ 55.1 / 54.7 / 55.2} & \textcolor{gray}{38.5/ 45.4 / 40.8 / 45.8} & \textcolor{gray}{35.8 / 42.9 / 38.3 / 43.2} \\\midrule
        \textbf{LoRA-F} & 74.1 / 76.9 / 75.5 / 79.0 & 34.4 / 41.4 / 37.7 / 42.9 & 24.2 / 30.8 / 27.2 / 32.7& 22.6 / 29.1 / 25.9 / 30.6 \\
        \textbf{TIES} &  40.0 / 42.5 / 41.1 / 44.4& 26.9 / 33.2 / 29.6 / 34.7& 13.3 / 18.4 / 15.8 / 19.9& 9.2 / 14.3 / 11.4 / 15.8 \\
        \textbf{DARE} &  36.2 / 38.9 / 37.4 / 40.7& 30.1 / 35.3 / 32.3 / 36.7& 12.0 / 17.3 / 14.5 / 18.8& 8.6 / 13.6 / 11.1 / 14.7 \\
        \textbf{Arrow} &  60.8 / 63.6 / 62.2 / 66.0& 23.0 / 28.5 / 25.7 / 29.9& 20.9 / 26.9 / 23.6 / 28.3& 16.9 / 22.9 / 19.9 / 24.2 \\
        \textbf{LoraHub} &  18.5 / 21.7 / 19.5 / 22.9& 6.5 / 9.6 / 8.1 / 10.1& 8.1 / 12.8 / 10.4 / 14.1& 5.8 / 9.4 / 7.4 / 10.4 \\
        \textbf{MeteoRA (T1-1k)} & 77.8 / 81.6 / 79.0 / 84.1& 37.4 / 43.6 / 40.5 / 45.6& 25.4 / 32.0 / 28.0 / 33.8& 24.0 / 29.7 / 26.6 / 31.4 \\\midrule
        \textbf{DLP-LoRA} &  \textbf{89.7} / \textbf{92.9} / \textbf{90.0} / \textbf{95.0}& \textbf{51.9} / \textbf{54.8} / \textbf{54.9} / \textbf{54.9}& \textbf{40.1} / \textbf{45.4} / \textbf{41.8} / \textbf{46.6}& \textbf{36.9} / \textbf{43.1} / \textbf{39.1} / \textbf{44.0} \\
        \bottomrule
    \end{tabular}}
    \caption{Average performance of 26 tasks on four different LLM backbones by comparing different LoRA baselines and our DLP-LoRA. For each column under the corresponding evaluation metric, the results represent Qwen-2 1.5B / Qwen-2 7B / LLaMA-2 7B / LLaMA-3 8B backbones used for each baseline, respectively. Our DLP-LoRA significantly outperforms all LoRA baselines across all tasks based on the average evaluation metric. For each task result, please refer to the Appendix~\ref{app:all_results}.}
    \label{table:all_baseline}
    \vspace{-1em}
\end{table*}

\paragraph{Datasets.}

To comprehensively evaluate our proposed DLP-LoRA framework, we follow the methodology of \citet{xu2025meteora} and conduct experiments across 26 diverse tasks. These include 18 multiple-choice question (MCQ) datasets covering domains such as mathematical question answering, logical reasoning, language identification, and reading comprehension. Additionally, we assess performance on 8 question-answering (QA) datasets focused on summarisation, machine translation, and open-domain QA. Specifically, we utilise 20 tasks from the BigBench benchmark~\citep{srivastava2023beyond}, 3 machine translation tasks from the News Commentary dataset~\citep{tiedemann2012parallel} translating from non-English to English, and 3 generative tasks: GSM8K~\citep{cobbe2021training}, CNN/DailyMail~\citep{see2017get}, and Alpaca~\citep{taori2023stanford}. Detailed descriptions of each dataset are provided in Appendix~\ref{app:26_datasets}.

\paragraph{LLM Backbones, LoRAs, and Mini-MLP Plugin.}

We compared DLP-LoRA with several LoRA baselines, such as TIES~\cite{yadav2024ties}, DARE~\cite{yu2024language}, Arrow~\cite{ostapenko2024towards}, LoraHub~\cite{huang2024lorahub} and MeteoRA (T1-1k)~\cite{xu2025meteora}, using four widely adopted LLM backbones: Qwen-2 1.5B and 7B~\citep{qwen2}, LLaMA-2 7B~\citep{touvron2023llama}, and LLaMA-3 8B~\citep{dubey2024llama}. In addition, we use Huggingface PEFT (i.e., PEFT) with all 26 LoRA loaded and manual activation for specific LoRA during evaluation as a reference model. We further train a single LoRA (i.e., LoRA-F) with a mixed training dataset from all 26 tasks for comparison.

For the baseline comparisons involving single LoRA modules, we fine-tune a separate LoRA for each task using 900 training samples, randomly selected according to a 9:1 train/test split from each original dataset following~\citep{xu2025meteora}. The rank of each LoRA used in baselines and our DLP-LoRA is 8. The mini-MLP plugin, responsible for task classification, is trained on the same samples and achieves an average classification accuracy of 98.45\%. Notably, the mini-MLP plugin is lightweight, containing only 5M parameters, and can be trained rapidly in under 10 minutes for all 26 tasks and easy to extend to 100 tasks without further fine-tuning the gating networks contained in MoE-structure baselines, such as MeteoRA. All experiments regarding DLP-LoRA and other baselines are conducted on a single NVIDIA GTX 3090Ti GPU 24GB and H100, respectively.

\paragraph{Evaluation Metrics and Composite Task Setting.}

Given that all 26 tasks can be categorised into MCQ and QA types, we employ accuracy as the evaluation metric for MCQ tasks and BLEU, ROUGE-1, and ROUGE-L scores for QA tasks. To assess multi-task learning capabilities, we create composite task settings by combining the 18 MCQ tasks (Composite-18) and the 8 QA tasks (Composite-8). In all experiments, we report the average results over 10 runs to ensure statistical reliability.

\subsection{Experimental Results}\label{sec:exp_results}

\paragraph{Main Results.}

Figure~\ref{fig:evaluation_26_llama2_7B} presents the performance of our DLP-LoRA compared to 7 LoRA baselines across 26 tasks using Qwen-2 1.5B and LLaMA-3 8B as backbones. Our DLP-LoRA not only significantly outperforms most LoRA baselines but also achieves performance comparable to, and in some cases surpassing, that of the manually loaded PEFT method across 26 tasks. Similar trends are observed for another two LLM backbones in Appendix~\ref{app:all_results}. As shown in Table~\ref{table:all_baseline}, DLP-LoRA achieves significant improvement on accuracy, BLEU, ROUGE-1 and ROUGE-L with the average 91.9\%, 54.1, 43.5 and 40.8 compared to SOTA MeteoRA, respectively. In addition, DLP-LoRA has comparable or better performance on MCQ tasks or QA tasks when using Qwen-2 7B, LLaMA-2 7B and LLaMA-3 8B than the PEFT reference approach. These results demonstrate that DLP-LoRA can match or even exceed the performance of individually fine-tuned single LoRAs or dynamic MoE-based LoRA baselines by dynamically selecting and fusing multiple LoRAs on the sentence level.

\begin{table}[tb]
    \centering
    \small
    \resizebox{\columnwidth}{!}{
    \begin{tabular}{llccc}
    \toprule
      \textbf{Composite-$n$} & \textbf{Metric (Avg.)} $\uparrow$   & \textbf{Basic} & \textbf{LoRA-F} ($r$=64) & \textbf{DLP-LoRA} \\\midrule
        \textbf{Composite-$18$}& Acc.   & 48.0& 49.1& \textbf{92.6}\\\midrule
       \multirow{3}{*}{\textbf{Composite-$8$}} &BLEU  & 52.3& 52.6& \textbf{57.5}\\
      & ROUGE-1  & 49.1& 49.5& \textbf{55.9}\\
       &ROUGE-L  & 46.5& 46.9& \textbf{53.8}\\
       \bottomrule
    \end{tabular}}
    \caption{Evaluation results for composite-$n$ task, where composite-8 includes all QA tasks, and composite-18 includes all MCQ tasks. In addition, we compare a single LoRA with a higher rank trained on composite-26 task setting. The evaluation results are averaged after running 10 times.}
    \label{ab:composite_n}
\end{table}

\paragraph{Multi-task Composite Performance.}

We further evaluate DLP-LoRA's capability in multi-task learning under composite task settings by combining the 18 MCQ tasks and the 8 QA tasks. As presented in Table~\ref{ab:composite_n}, DLP-LoRA significantly enhances performance over the basic LLM backbones, achieving absolute improvements of 44.6\% in accuracy for the MCQ composite, and 5.2, 6.8, and 7.3 in BLEU, ROUGE-1, and ROUGE-L scores, respectively, for the QA composite. In addition, we further fine-tuned a single LoRA with a higher rank 64 on all 26 tasks, and the improvement of such LoRA-F ($r=64$) is incremental, which confirmed the argument that a single adapter on a combined dataset can dilute domain-specific knowledge~\citep{lin2024teamlora}. These findings indicate that DLP-LoRA effectively and automatically selects the appropriate LoRAs based on the input prompts within composite tasks, facilitating dynamic multi-task adaptation. A detailed example illustrating how DLP-LoRA selects and fuses multiple LoRAs is provided in Section~\ref{sec:case_study}.

\paragraph{Inference Time Efficiency Compared to LLM Backbones.}

We also conduct a comprehensive evaluation of the inference time efficiency of DLP-LoRA and its variants compared to the basic LLM backbones and single LoRA models. As shown in Table~\ref{tab:inference_time_rate}, single LoRA models exhibit inference speeds comparable to the baseline LLMs, being only about 1.05x slower on average. When incorporating ALBERT (11M parameters) as the plugin, DLP-LoRA's inference time ranges from 1.12 to 1.90x slower than the basic LLMs, representing a 41.90\% increase compared to single LoRA inference. By contrast, using the mini-MLP plugin with 5M parameters, DLP-LoRA achieves faster inference, with only an 18.10\% average increase in inference time over single LoRA models across all tasks. These results validate the efficiency of our sentence-level LoRA selection and fusion approach.

\begin{table}[tb]
    \centering
    \small
    \resizebox{\columnwidth}{!}{
    \begin{tabular}{lccc}
    \toprule
       \textbf{Models}  & \textbf{LoRA}& \textbf{DLP (ALBERT)} & \textbf{DLP (mini-MLP)} \\\midrule
        Qwen-2 1.5B & 1.15& ${{1.90}_{\textcolor{teal}{+65.22\%}}}$& $\bm{{1.12}_{\textcolor{teal}{-2.61\%}}}$\\
        Qwen-2 7B &1.00& ${{1.13}_{\textcolor{teal}{+13.00\%}}}$&$\bm{{1.12}_{\textcolor{teal}{+12.00\%}}}$\\
        LLaMA-2 7B &1.05& ${{1.80}_{\textcolor{teal}{+71.43\%}}}$&$\bm{{1.60}_{\textcolor{teal}{+52.38\%}}}$\\
        LLaMA-3 8B &1.00& ${{1.12}_{\textcolor{teal}{+12.00\%}}}$&$\bm{{1.11}_{\textcolor{teal}{+11.00\%}}}$\\\midrule
        Avg. & 1.05 & ${1.49}_{\textcolor{teal}{+41.90\%}}$ & $\bm{{1.24}_{\textcolor{teal}{+18.10\%}}}$\\
        \bottomrule
    \end{tabular}}
    \caption{The averaged inference time ratio across 26 datasets by comparing the single LoRA, and DLP-LoRA equipped ALBERT and mini-MLP plugin with the basic LLMs backbones. The subscript percentage denotes relative inference time improvement or reduction of DLP-LoRA over the single LoRA inference.}
    \label{tab:inference_time_rate}
\end{table}

\paragraph{Efficiency Comparison among Different Dynamic LoRAs Baselines.}
\begin{table}[tb]
    \centering
    \small
    \resizebox{\columnwidth}{!}{
    \begin{tabular}{lccc}
    \toprule
      \textbf{Models}   &  \textbf{Decoding latency ratio} & \textbf{Peak Memory ratio} \\\midrule
       \textbf{LLaMA2-7B}  & 1.00 &  1.00\\\midrule
       \textbf{MOLA} & ${10.54}_{\textcolor{teal}{+954\%}}$ &  ${2.04}_{\textcolor{teal}{+104\%}}$ \\
       \textbf{PESC} & ${3.54}_{\textcolor{teal}{+254\%}}$ &  ${{{1.02}}_{\textcolor{teal}{+2\%}}}$ \\
       \textbf{MoRAL}  & ${3.58}_{\textcolor{teal}{+258\%}}$ &  ${{{1.02}}_{\textcolor{teal}{+2\%}}}$ \\
       \textbf{LoRA-Switch} & ${{1.29}_{\textcolor{teal}{+29\%}}}$ &  ${1.07}_{\textcolor{teal}{+7\%}}$ \\
       \textbf{MeteoRA} & ${2.62}_{\textcolor{teal}{+162\%}}$ & ${1.12}_{\textcolor{teal}{+12\%}}$\\\midrule
       \textbf{DLP-LoRA} & $\bm{{{1.20}}_{\textcolor{teal}{+20\%}}}$ &  $\bm{{1.00}_{\textcolor{teal}{+0\%}}}$ \\
    \bottomrule
    \end{tabular}}
    \caption{The inference time and memory consuming ratio compared with different dynamic LoRAs baselines based on LLaMA-2 7B. The subscript percentage denotes relative inference time and memory usage improvement of different LoRAs baselines over the LLaMA-2 7B backbone.}
    \label{tab:ablation_method}
\end{table}
\begin{table*}[tb]
    \centering
    \small
    \resizebox{\textwidth}{!}{
    \begin{tabular}{p{18cm}|c}
    \toprule
    Content & Contributions of LoRAs \\\midrule
       \textbf{Query:} Is it common to see frost during some college commencements? & AbsNarr: 50.5\%\\
       \textbf{Output:} Yes, it is common to see frost during some college commencements, especially in areas where the weather is cooler or more temperate. Frost is the formation of ice crystals on surfaces in the atmosphere, typically when the temperature is below freezing. It can occur on surfaces such as grass, leaves, and even clothing.

In some cases, frost can form on the ground or on surfaces during college commencements, especially if the temperature is around or below freezing. This can happen in the early morning or late evening, when the air is colder.
  & GSM8K: 49.5\%\\\midrule
  \textbf{Query:} Aber das Ausmaß dieser Korruption sollte uns angesichts der Probleme mit der Buchführung der öffentlichen Hand nicht blenden. Auch in diesem Bereich werden hinterhältige Dinge gedreht. & NewsDE: 100.0\%\\
  \textbf{Output:} But the freezing of this corruption should not blind us to the problems with the public sectors accounting. In this area, too, something is being covered up. & \\\midrule
  \textbf{Query:} It is not always easy to see which chemicals are contained in our consumer products. The following argument pertains to this question: First premise: Selenium sulfide is an ingredient of Water Baby. Second premise: Being an ingredient of Climate Control is sufficient for not being an ingredient of Water Baby. We may conclude that selenium sulfide is not an ingredient of Climate Control. Is the argument, given the explicitly stated premises, deductively valid or invalid? A valid B invalid: & FormFall: 100.0\%\\
  \textbf{Output: A} & \\
  \bottomrule

    \end{tabular}}
    \caption{Case study of DLP-LoRA based on LLaMA-3 8B backbone under composite-3 task setting, where the selected LoRAs with corresponding probabilities are demonstrated on the right side.}
    \label{tab:case_study}
\end{table*}
We further compare our DLP-LoRA with different dynamic LoRAs baselines in order to evaluate the DLP-LoRA's efficiency at inference speed and memory usage. We fine-tuned 8 different LoRAs based on selected 8 datasets including ARC~\citep{clark2018think}, HellaSwag~\citep{zellers2019hellaswag}, MMLU~\citep{hendrycks2020measuring}, TruthfulQA~\citep{lin2022truthfulqa}, WinoGrande~\citep{sakaguchi2021winogrande}, ScienceQA~\citep{lu2022learn}, CommonsenseQA~\citep{talmor2019commonsenseqa}, and OpenbookQA~\citep{mihaylov2018can}. Then we compare DLP-LoRA with different baselines, such as MOLA~\cite{gao2024higher}, PESC~\cite{wu2024parameter}, MoRAL~\cite{yang2024moral} and LoRA-Switch~\cite{kong2024lora} on the ShareGPT dataset~\citep{wang2023openchat}~\footnote{Since LoRA-Switch did not descript how many LoRAs are utilised during inference for ShareGPT dataset, we assume that all 8 LoRAs based on the original work are equipped 
 and we can regard this as the lower-bound of DLP-LoRA.}. As shown in Table~\ref{tab:ablation_method}, it is evident that DLP-LoRA stands out in both speed and memory efficiency. Even when handling 8 tasks, DLP-LoRA completes inference tasks fast with only 1.20x slower than the basic LLaMA-2 7B inference and with minimal additional memory costs, demonstrating a significant advantage over other dynamic LoRA baselines.

\subsection{Case Study}\label{sec:case_study}

To illustrate the practical effectiveness of DLP-LoRA, we present a case study in Table~\ref{tab:case_study} using the LLaMA-3 8B backbone under a composite task setting involving three tasks. For the first input prompt, DLP-LoRA selects two LoRAs, i.e., AbsNarr and GSM8K, with probabilities of 50.5\% and 49.5\%, respectively, using top-$p$ sampling. The AbsNarr dataset involves narratives encapsulating human experiences and wisdom, while GSM8K focuses on practical scenarios requiring general knowledge through mathematical reasoning. The gold standard dataset, StratQA, requires answering general knowledge questions with reasoning steps. DLP-LoRA effectively fuses the AbsNarr and GSM8K LoRAs to generate logical explanations that incorporate general knowledge about frost weather and commencements. When subsequent questions are input, concatenated with the history, DLP-LoRA continues to successfully select the appropriate LoRAs, i.e., NewsDE and FormFall, from the pool of 26 LoRAs stored in high-bandwidth memory (HBM). This case study demonstrates DLP-LoRA's ability to dynamically select and fuse multiple LoRAs to address diverse tasks effectively.

%% file: sections/discussion.tex
\section{Discussion}
\begin{table*}[tb]
    \centering
    \small
    \resizebox{\textwidth}{!}{
    \begin{tabular}{lccc}
    \toprule
    \textbf{Models} & \textbf{Param.} (layer-wise) &  \textbf{Fusion} & \textbf{Dynamic}\\\midrule
       \textcolor{gray}{\textbf{PEFT (Ref.)}}  & $2(A+B)$ & \xmark& \xmark\\\midrule
        \textbf{LoRA-F} & $2(A+B)$& Manually merge all datasets&\xmark\\
        \textbf{TIES} &$2(A+B)$& Trim redundancy + Merge aligned vectors&\xmark\\
        \textbf{DARE} &$2(A+B)$&Random drop + Rescale delta parameters + Merge&\xmark\\
        \textbf{Arrow} &$2(hN+A+B)$&SVD of each LoRA params from built Model-Based Clustering LoRAs&\cmark\\
        \textbf{LoraHub} &$2(A+B)$&Compose multiple LoRAs + Adapt the set of coefficients based evolution strategies&\xmark\\
        \textbf{MeteoRA (T1-1k)} &$7(hN+A+B)$&Token-level trainable Gating network added to 7 modules per layer&\cmark\\\midrule
        \textbf{DLP-LoRA} &$\frac{5e6}{L}+2(A+B)$&A 5M mini-MLP plugin to dynamically fuse multiple LoRAs&\cmark\\
        \bottomrule
    \end{tabular}}
    \caption{The layer-wise LoRA parameters comparison among different baselines and our DLP-LoRA with corresponding LoRA fusion methods, where $A, B, h, N, L$ indicate the parameters of LoRA's A, B matrices, model's hidden representations, number of LoRAs and number of total layers, respectively. Apart from MeteoRA which is designed to add a gating network with LoRA to 7 components per layer, other LoRA baselines and our DLP-LoRA only introduce LoRAs to the query and value projections in the attention layer.}
    \label{tab:all_baseline_param}
\end{table*}
\paragraph{Limitations of Top-$k$ Selection.}

Most existing Multi-LoRA or LoRA-MoE methods employ a top-$k$ router to manually determine the fixed number of LoRAs to use for multi-task learning~\citep{li2024mixlora, yang2024moral, wu2024parameter}. This manual selection can restrict the model's ability to dynamically select and fuse multiple LoRAs based on the task requirements. In our approach, we utilise top-$p$ selection, which leverages the probabilities assigned by the mini-MLP plugin to each LoRA, using a threshold $p$. This allows DLP-LoRA to adaptively decide both the number and combination of LoRAs to fuse for different tasks, enhancing flexibility and performance.

\paragraph{Additional Parameters Added by Different LoRAs.}
Apart from the performance comparison in Table~\ref{table:all_baseline}, we further analyse how many additional parameters are introduced for each LoRA baseline compared to our DLP-LoRA in Table~\ref{tab:all_baseline_param}. We demonstrate the layer-wise parameters added to the LLM backbones, and indicate the fusion strategy and whether each LoRA baseline is dynamic. As demonstrated in Table~\ref{tab:all_baseline_param}, DLP-LoRA only introduces $\frac{5e6}{L}$ parameters\footnote{Those new introduced parameters are the mini-MLP, and it accounts for 5M in total when we sum up across all layers.} per layer compared to all static LoRA baselines. When compared to other two dynamic LoRA baselines, i.e., Arrow and MeteoRA, our DLP-LoRA has a superior advantage, as Arrow has to implement SVD decomposition for all LoRAs to build layer-wise weight matrices for hidden states routing and MeteoRA inserts the trainable gating network with MoE on 7 components (Q, K, V and O in the attention layer and up-projection, gating for SiLU and down-projection in MLP) per layer.

\begin{table}[tb]
    \centering
    \resizebox{\columnwidth}{!}{
    \begin{tabular}{lccc}
    \toprule
     \textbf{Models} & \textbf{Num. of LoRA}   & \textbf{\# Params} (\%) & \textbf{Inference Time Ratio} \\\midrule
     \multirow{2}{*}{MeteoRA (T1-1k)} & 50 & 2.065& 3.75\\
     & 100 & 8.483& 4.02\\\midrule
      \multirow{2}{*}{DLP-LoRA} &50  & 0.043&  1.76\\
       &100 & 0.085&  1.83 \\
    \bottomrule
    \end{tabular}}
    \caption{The increased LoRA's parameters and inference time ratio compared between MeteoRA (T1-1k) and our DLP-LoRA under different numbers of LoRAs using the LLaMA-3 8B as the backbone. \# Params denote the percentage of LoRAs' parameters over the LLaMA-3 8B.}
    \label{tab:ablation_lora}
\end{table}

\paragraph{Inference Time of Multi-LoRA Loading at Scale}
Table~\ref{table:all_baseline} shows the superior performance of our DLP-LoRA compared to other LoRA baselines across 26 tasks. It is also important to demonstrate whether the inference time is practical when more LoRAs are required in real-world settings.
We conducted an ablation study to assess how the inference time scales with the increasing number of LoRAs, using the LLaMA-3 8B backbone as a reference. As illustrated in Table~\ref{tab:ablation_lora}, even as the number of LoRAs increases to 100, the inference time ratio of DLP-LoRA remains within 2x using the LLaMA-3 8B model. Additionally, the combined parameters of all LoRAs constitute less than 0.1\% of the 8B parameters in the LLaMA-3 backbone. With our DLP plugin method, switching to a different LoRA requires only retraining a small 5M mini-MLP, resulting in minimal computational overhead. However, MeteoRA needs to further insert and fine-tune the whole seven trainable gating networks per layer for all introduced new LoRAs, which significantly increases the number of new parameters and computational resources. In contrast, DLP-LoRA only adjusts the final linear layer of mini-MLP, which keeps the total increase to around 5M parameters. This suggests that LoRA fine-tuning can enable LLMs to enhance their capabilities across various domains simultaneously when equipped with sufficient LoRAs. In summary, these results in Table~\ref{tab:ablation_lora} demonstrate that our approach scales efficiently with the number of LoRAs without incurring significant computational overhead, maintaining practical inference times even at scale.

%% file: sections/related_work.tex
\section{Related Work}

In the area of multi-task learning with LoRA, two primary research directions have emerged beyond the straightforward approach of fine-tuning a single LoRA on a combined dataset of multiple tasks~\citep{lin2024teamlora}. The first direction focuses on developing libraries or frameworks to reuse and integrate existing LoRAs, while the second aims to design router networks based on MoEs to dynamically fuse multiple LoRAs.

\paragraph{Multiple LoRA Architectures}

Several works have proposed frameworks for combining and managing multiple LoRAs. \citet{huang2023lorahub} introduced LoRAHub, a framework that combines existing fine-tuned LoRAs using a learnable weighted sum, allowing for more flexible adaptation across tasks. S-LoRA~\citep{sheng2023s} emphasises unified memory pool design to manage dynamic LoRA weights with varying ranks and key-value cache tensors for CUDA kernels, enhancing computational efficiency. Additionally, Model-Based Clustering (MBC)~\citep{ostapenko2024towards} employs clustering techniques to group tasks based on the similarity of their LoRA parameters, facilitating better parameter sharing and task generalization.

\paragraph{Mixture-of-Experts with Multiple LoRAs}

Another line of research integrates Mixture-of-Experts mechanisms to control and fuse multiple LoRAs dynamically. In these approaches, multiple LoRAs are fine-tuned and injected into the model's MLP layers, with a router network determining which LoRA to activate for a given task. Examples include LoRAMoE~\citep{dou2024loramoe}, PHATGOOSE~\citep{muqeeth2024learning}, MoLE~\citep{wu2024mixture}, and LoRA-Switch~\citep{kong2024lora}. Some methods extend this fusion to both MLP and attention layers, such as MixLoRA~\citep{li2024mixlora} and Mixture of Adaptations (MoA)~\citep{feng2024mixture}, enabling more comprehensive adaptation across model components.

Furthermore, token-level routing strategies have been proposed to enhance the granularity of LoRA selection. MeteoRA~\citep{xu2025meteora} introduces a token-level MoE-style multi-task LoRA framework with trainable gating mechanisms across all attention and MLP layers, allowing for dynamic selection and fusion of different LoRAs based on input tokens. Similarly, AdaMoE~\citep{zeng2024adamoe} presents an adaptive MoE approach that leverages token-level routing within transformer models to improve performance across diverse tasks.

%% file: sections/appendix.tex
\section{Broader Impacts}

The lightweight design of DLP-LoRA, featuring a mini-MLP with only 5 million parameters, offers significant flexibility and efficiency, making it suitable for deployment on smaller devices with limited computational resources. Moreover, DLP-LoRA facilitates easy integration of new LoRAs corresponding to additional tasks without necessitating further fine-tuning of the entire model. This capability enhances the accessibility and adaptability of LLMs in various applications, promoting broader utilisation in resource-constrained environments.

\section{Details about 26 Tasks and Datasets}\label{app:26_datasets}
Table~\ref{tab:28_tasks_details} includes detailed descriptions of each dataset's name, keywords, main content and corresponding evaluation metrics. These 26 tasks include diverse topics, such as mathematical QA, logical reasoning, language identification, reading comprehension, summarisation, machine translation, and open-domain QA.

\begin{table*}[!h]
\centering
\renewcommand{\arraystretch}{1.4}
\resizebox{\textwidth}{!}{
\begin{tabular}{cp{3.5cm}p{7cm}p{2.8cm}}
\toprule
Task Name & Keywords & Description & Evaluation Metrics \\ 
\midrule
abstract\_narrative\_understanding (AbsNarr) &
  narrative understanding, multiple choice &
  Given a narrative, choose the most related proverb. &
  Accuracy \\
alpaca (ALPACA) &
  instruction-tuning &
  Write appropriate answers according to instructions. &
  BLEU, ROUGE \\
cnn\_dailymail (CNNDM) &
  summarization &
  Given news articles, write the summarization. &
  ROUGE \\
contextual\_parametric\_knowledge\_conflicts (ConParaKC) &
  contextual question-answering, multiple choice &
  Answer questions given the contextual information. &
  Accuracy \\
cs\_algorithms (CSAlg) &
  algorithms, numerical response &
  Solve two common computer-science tasks. &
  Accuracy \\
disfl\_qa (DisflQA) &
  contextual question-answering, reading comprehension &
  Pick the correct answer span from the context given the disfluent question. &
  Accuracy \\
elementary\_math\_qa (ElemMath) &
  mathematics &
  Answer multiple choice mathematical word problems. &
  Accuracy \\
epistemic\_reasoning (EpiReason) &
  logical reasoning, multiple choice &
  Determine whether one sentence entails the next. &
  Accuracy \\
formal\_fallacies\_syllogisms\_negation (FormFall) &
  logical reasoning, multiple choice, &
  Distinguish deductively valid arguments from formal fallacies. &
  Accuracy \\
gsm8k (GSM8K) &
  mathematics &
  Solve the grade school math word problems. &
  Accuracy \\
language\_identification (LangID) &
  multilingual, multiple choice &
  Given a sentence, select the correct language. &
  Accuracy \\
linguistics\_puzzles (LingPuzz) &
  logical reasoning, linguistics &
  Solve Rosetta Stone-style linguistics puzzles. &
  BLEU, ROUGE \\
logical\_deduction (LogDeduc) &
  logical reasoning, multiple choice &
  Deduce the order of a sequence of objects. &
  Accuracy \\
news\_commentary\_de (NewsDE) &
  multilingual, translation &
  Translate German sentences into English. &
  BLEU \\
news\_commentary\_es (NewsES) &
  multilingual, translation &
  Translate Spanish sentences into English. &
  BLEU \\
news\_commentary\_it (NewsIT) &
  multilingual, translation &
  Translate Italian sentences into English. &
  BLEU \\
object\_counting (ObjCount) &
  logical reasoning &
  Questions that involve enumerating objects and asking the model to count them. &
  Accuracy \\
play\_dialog\_same\_or\_different (PlayDiag) &
  reading comprehension, multiple choice &
  Determine if nearby lines in a Shakespeare play were spoken by the same individual. &
  Accuracy \\
question\_selection (QuestSel) &
  reading comprehension, multiple choice &
  Given an answer along with its context, select the most appropriate question which has the given answer as its answer. &
  Accuracy \\
reasoning\_about\_colored\_objects (ColorReason) &
  reading comprehension, logical reasoning, multiple choice &
  Answer extremely simple questions about the colors of objects on a surface. &
  Accuracy \\
strategyqa (StratQA) &
  logical reasoning, context-free question answering &
  Answer questions in which the required reasoning steps are implicit in the question. &
  BLEU, ROUGE, Accuracy \\
topical\_chat (TopChat) &
  free response &
  Open-domain response generation. &
  BLEU, ROUGE \\
tracking\_shuffled\_objects (TrackObj) &
  logical reasoning, multiple choice &
  Determine the final positions given initial positions and a description of a sequence of swaps. &
  Accuracy \\
unit\_conversion (UnitConv) &
  contextual question-answering, mathematics, multiple choice &
  Perform various tasks relating to units, including identification and conversion. &
  Accuracy \\
vitaminc\_fact\_verification (VitaFact) &
  truthfulness, reading comprehension, multiple choice &
  Identify whether a claim is True or False based on the given context. &
  Accuracy \\
winowhy (WinoWhy) &
  causal reasoning, multiple choice &
  Evaluate the reasoning in answering Winograd Schema Challenge questions. &
  Accuracy \\ 
\bottomrule
\end{tabular}}
\caption{Details about the 26 selected tasks following~\cite{xu2025meteora}.}
\label{tab:28_tasks_details}
\end{table*}

\section{Experimental Results on All Datasets}\label{app:all_results}
Table~\ref{table:acc_all_results_qwen_2_1.5b},~\ref{table:acc_all_results_qwen_2_7b},~\ref{table:acc_all_results_llama_2_7b} and~\ref{table:acc_all_results_llama_3_8B} show all results among different LoRA baselines and DLP-LoRA using Qwen-2 1.5B, Qwen-2 7B, LLaMA-2 7B and LLaMA-3 8B backbones.
\begin{table*}[tb]
    \centering
    \resizebox{\textwidth}{!}{
    \begin{tabular}{lc|cccccc|c}
        \toprule
        \textbf{Models} & \textbf{PEFT (Ref.)} & \textbf{LoRA-F} & \textbf{TIES} & \textbf{DARE} & \textbf{Arrow} & \textbf{LoraHub} & \textbf{MeteoRA (T1-1k)} & \textbf{DLP-LoRA}\\
        \midrule
        \textbf{AbsNarr}   &89.3&75.6&36.8&28.9&72.2&8.1&71.6&89.8\\
        \textbf{ConParaKC} &100.0&94.0&55.9&43.9&92.0&17.5&91.8&93.8\\
        \textbf{CSAlg}     &97.5&76.9&62.7&63.0&72.6&54.0&76.1&98.8\\
        \textbf{DisflQA}   &87.6&65.9&33.3&32.1&44.9&13.7&65.4&88.1\\
        \textbf{ElemMath}  &81.0&74.8&25.4&20.0&55.9&7.8&73.9&81.3\\
        \textbf{EpiReason} &99.8&97.2&41.7&37.8&54.6&13.5&98.0&99.5\\
        \textbf{FormFall}  &100.0&93.5&46.6&47.5&75.8&15.4&94.1&100.0\\
        \textbf{GSM8K}&86.0&50.2&5.0&8.6&13.2&2.6&48.5&85.3\\
        \textbf{LangID}    &77.0&67.6&37.0&33.7&56.9&14.8&71.4&77.0\\
        \textbf{LogDeduc}  &84.5&65.4&38.6&31.5&59.8&30.6&68.9&80.8\\
        \textbf{ObjCount}  &89.0&51.8&10.0&12.7&5.4&17.6&80.6&88.0\\
        \textbf{PlayDiag}  &89.0&56.3&58.7&57.6&59.3&6.0&57.4&88.0\\
        \textbf{QuesSel}   &99.0&85.4&45.6&41.9&78.4&19.4&86.7&98.0\\
        \textbf{ColorReason} &79.0&88.0&54.2&48.6&72.7&18.6&90.5&78.3\\
        \textbf{TrackObj}  &79.8&54.8&14.6&11.0&74.0&13.2&90.3&78.8\\
        \textbf{UnitConv}  &100.0&75.4&44.8&36.9&58.9&39.4&74.7&100.0\\
        \textbf{VitaFact}  &94.0&86.2&60.6&50.1&77.6&26.8&85.8&92.3\\
        \textbf{WinoWhy}   &94.8&74.6&48.0&46.4&70.7&14.6&75.2&96.0\\\midrule
        \textbf{Avg.}&90.4&74.1&40.0&36.2&60.8&18.5&77.8&89.7\\
        \bottomrule
    \end{tabular}}
    \caption{The classification accuracy results on 18 MCQ tasks by comparing different LoRA baselines under Qwen-2 1.5B as LLM backbone. The evaluation results are averaged after running 10 times.}
    \label{table:acc_all_results_qwen_2_1.5b}
\end{table*}

\begin{table*}[tb]
    \centering
    \resizebox{\textwidth}{!}{
    \begin{tabular}{lc|cccccc|c}
        \toprule
        \textbf{Models} & \textbf{PEFT (Ref.)} & \textbf{LoRA-F} & \textbf{TIES} & \textbf{DARE} & \textbf{Arrow} & \textbf{LoraHub} & \textbf{MeteoRA (T1-1k)} & \textbf{DLP-LoRA}\\
        \midrule
        \textbf{AbsNarr}   &93.3&78.5&40.2&30.2&75.6&10.4&76.2&92.8\\
        \textbf{ConParaKC} &99.0&96.1&57.7&47.8&95.5&20.8&95.3&94.0\\
        \textbf{CSAlg}     &100.0&80.8&65.2&64.4&74.7&57.9&81.3&100.0\\
        \textbf{DisflQA}   &89.6&67.2&36.8&35.7&47.8&18.1&68.5&88.0\\
        \textbf{ElemMath}  &85.8&76.8&28.6&22.5&60.3&8.9&76.4&86.0\\
        \textbf{EpiReason} &100.0&99.2&43.6&40.8&57.3&16.4&99.4&100.0\\
        \textbf{FormFall}  &100.0&95.2&49.7&49.5&78.9&18.1&96.2&100.0\\
        \textbf{GSM8K}&93.4&55.3&7.7&10.0&16.2&4.8&53.9&93.3\\
        \textbf{LangID}    &89.3&71.6&39.8&36.2&59.9&18.4&75.8&88.0\\
        \textbf{LogDeduc}  &89.5&67.9&40.5&35.8&61.2&34.7&73.2&90.8\\
        \textbf{ObjCount}  &94.7&53.6&8.6&13.4&2.4&21.7&82.8&93.9\\
        \textbf{PlayDiag}  &90.8&59.5&62.4&61.9&62.7&8.1&60.3&89.8\\
        \textbf{QuesSel}   &98.0&88.7&48.2&45.7&81.4&22.7&90.9&97.0\\
        \textbf{ColorReason} &87.5&92.5&57.6&51.3&76.8&21.5&95.3&87.8\\
        \textbf{TrackObj}  &81.0&56.8&17.4&12.1&77.9&15.6&97.4&82.3\\
        \textbf{UnitConv}  &100.0&78.9&47.3&39.5&62.3&43.7&79.6&100.0\\
        \textbf{VitaFact}  &96.5&87.8&63.6&52.5&80.3&29.8&88.4&95.5\\
        \textbf{WinoWhy}   &91.3&77.4&49.4&50.2&73.7&18.4&78.5&93.5\\\midrule
        \textbf{Avg.} &93.3&76.9&42.5&38.9&63.6&21.7&81.6&92.9\\
        \bottomrule
    \end{tabular}}
    \caption{The classification accuracy results on 18 MCQ tasks by comparing different LoRA baselines under Qwen-2 7B as LLM backbone. The evaluation results are averaged after running 10 times.}
    \label{table:acc_all_results_qwen_2_7b}
\end{table*}

\begin{table*}[tb]
    \centering
    \resizebox{\textwidth}{!}{
    \begin{tabular}{lc|cccccc|c}
        \toprule
        \textbf{Models} & \textbf{PEFT (Ref.)} & \textbf{LoRA-F} & \textbf{TIES} & \textbf{DARE} & \textbf{Arrow} & \textbf{LoraHub} & \textbf{MeteoRA (T1-1k)} & \textbf{DLP-LoRA}\\
        \midrule
        \textbf{AbsNarr}   &92.5&76.8&38.9&29.4&73.9&8.5&72.8&89.5\\
        \textbf{ConParaKC} &96.0&95.4&56.7&45.8&93.2&18.4&92.4&92.8\\
        \textbf{CSAlg}     &99.0&78.6&64.2&63.2&74.3&54.7&77.5&98.8\\
        \textbf{DisflQA}   &89.0&66.8&34.9&33.8&45.6&15.8&66.4&91.2\\
        \textbf{ElemMath}  &78.0&75.7&26.7&21.2&58.5&7.7&74.7&80.0\\
        \textbf{EpiReason} &100.0&98.5&42.3&39.3&55.3&14.8&99.1&100.0\\
        \textbf{FormFall}  &100.0&94.2&48.2&48.2&77.5&16.3&95.4&100.0\\
        \textbf{GSM8K}&79.8&53.0&6.1&9.1&14.8&3.0&50.3&78.9\\
        \textbf{LangID}    &79.8&69.7&38.1&34.8&57.6&15.4&72.5&79.8\\
        \textbf{LogDeduc}  &83.0&66.8&39.4&33.9&60.4&31.9&70.0&82.8\\
        \textbf{ObjCount}  &91.1&52.6&9.4&13.2&4.6&19.9&81.5&90.7\\
        \textbf{PlayDiag}  &87.8&57.9&60.9&59.7&60.9&6.8&58.8&88.3\\
        \textbf{QuesSel}   &99.0&86.7&46.0&43.8&80.1&20.6&88.4&99.0\\
        \textbf{ColorReason} &80.8&90.4&55.8&49.6&74.9&19.0&91.4&80.8\\
        \textbf{TrackObj}  &80.0&55.6&15.8&11.5&75.4&13.8&92.7&78.8\\
        \textbf{UnitConv}  &100.0&76.9&45.7&37.8&60.9&40.9&75.9&100.0\\
        \textbf{VitaFact}  &90.9&87.0&61.9&51.9&79.6&27.3&86.5&92.7\\
        \textbf{WinoWhy}   &94.3&75.8&48.6&47.7&71.8&15.8&76.3&96.3\\\midrule
        \textbf{Avg.}&90.1&75.5&41.1&37.4&62.2&19.5&79.0&90.0\\
        \bottomrule
    \end{tabular}}
    \caption{The classification accuracy results on 18 MCQ tasks by comparing different LoRA baselines under LLaMA-2 7B as LLM backbone. The evaluation results are averaged after running 10 times.}
    \label{table:acc_all_results_llama_2_7b}
\end{table*}

\begin{table*}[tb]
    \centering
    \resizebox{\textwidth}{!}{
    \begin{tabular}{lc|cccccc|c}
        \toprule
        \textbf{Models} & \textbf{PEFT (Ref.)} &  \textbf{LoRA-F} & \textbf{TIES} & \textbf{DARE} & \textbf{Arrow} & \textbf{LoraHub} & \textbf{MeteoRA (T1-1k)} & \textbf{DLP-LoRA}\\
        \midrule
        \textbf{AbsNarr}  &97.4&79.3&42.5&33.5&77.2&7.5&78.7&97.3 \\
        \textbf{ConParaKC} &98.0&99.9&59.4&49.2&99.7&21.9&99.9&95.1\\
        \textbf{CSAlg}     &99.5&84.1&68.6&66.3&78.0&60.2&84.5&99.0\\
        \textbf{DisflQA}   &94.4&68.0&39.6&37.7&50.4&19.7&70.6&90.0\\
        \textbf{ElemMath}  &90.0&77.7&30.8&24.5&64.5&10.6&77.6&90.5\\
        \textbf{EpiReason} &100.0&99.6&45.0&42.5&60.0&17.0&100.0&100.0\\
        \textbf{FormFall}  &100.0&97.0&51.9&52.0&83.6&19.0&98.7&100.0\\
        \textbf{GSM8K}&81.6&56.6&8.6&10.8&17.2&5.0&55.5&79.1\\
        \textbf{LangID}    &95.1&74.9&41.2&38.3&62.5&19.2&77.9&94.5\\
        \textbf{LogDeduc}  &96.0&70.7&42.3&38.3&62.7&36.7&75.7&96.4\\
        \textbf{ObjCount}  &97.1&55.5&8.0&13.0&0.5&23.0&87.5&97.3\\
        \textbf{PlayDiag}  &95.0&63.2&65.0&64.4&65.6&9.2&64.9&94.8\\
        \textbf{QuesSel}   &97.0&91.1&50.6&47.2&84.5&24.7&92.7&97.0\\
        \textbf{ColorReason} &95.6&94.5&59.5&53.0&79.3&23.8&96.0&96.3\\
        \textbf{TrackObj}  &90.0&58.8&19.5&13.6&80.4&17.1&99.3&90.5\\
        \textbf{UnitConv}  &100.0&81.4&49.1&41.0&64.7&46.3&82.0&100.0\\
        \textbf{VitaFact}  &95.4&90.3&65.5&54.1&82.2&31.1&90.7&95.4\\
        \textbf{WinoWhy}   &96.9&79.7&51.6&52.6&75.0&20.3&81.8&96.9\\\midrule
        \textbf{Avg.} &95.5&79.0&44.4&40.7&66.0&22.9&84.1&95.0\\
        \bottomrule
    \end{tabular}}
    \caption{The classification accuracy results on 18 MCQ tasks by comparing different LoRA baselines under LLaMA-3 8B as LLM backbone. The evaluation results are averaged after running 10 times.}
    \label{table:acc_all_results_llama_3_8B}
\end{table*}

\begin{table*}[tb]
    \centering
    \resizebox{\textwidth}{!}{
    \begin{tabular}{lcc|cccccc|c}
        \toprule
        \textbf{Models} & \textbf{Metric} & \textbf{PEFT (Ref.)}  & \textbf{LoRA-F} & \textbf{TIES} & \textbf{DARE} & \textbf{Arrow} & \textbf{LoraHub} & \textbf{MeteoRA (T1-1k)} & \textbf{DLP-LoRA}\\
        \midrule
        
        \multirow{3}{*}{\textbf{CNNDM}}
            & BLEU     & 15.1 & 10.1 & 11.2 & 1.0 & 7.5 & 8.3 & 6.4 & 18.6 \\
            & ROUGE-1  & 16.9 & 17.2 & 7.6 & 5.8 & 7.9 & 2.3 & 13.2 & 19.0 \\
            & ROUGE-L  & 15.8 & 15.9 & 3.3 & 2.7 & 5.2 & 0.8 & 14.2 & 17.2 \\\midrule
        \multirow{3}{*}{\textbf{LingPuzz}}
            & BLEU     & 43.3 & 28.9 & 27.5 & 49.6 & 53.7 & 30.2 & 35.9 & 42.0 \\
            & ROUGE-1  & 29.4 & 55.7 & 33.8 & 26.9 & 61.2 & 17.9 & 60.2 & 26.7 \\
            & ROUGE-L  & 27.8 & 52.9 & 24.8 & 22.9 & 55.0 & 12.1 & 54.8 & 26.0 \\\midrule
        \multirow{3}{*}{\textbf{NewsDE}}
            & BLEU     & 64.2 & 65.8 & 36.4 & 28.6 & 28.9 & 5.0 & 74.8 & 64.3 \\
            & ROUGE-1  & - & - & - & - & - & - & - & - \\
            & ROUGE-L  & - & - & - & - & - & - & - & - \\\midrule
        \multirow{3}{*}{\textbf{NewsES}}
            & BLEU     & 66.7 & 68.9 & 20.3 & 9.4 & 21.9 & 0.0 & 71.8 & 67.3 \\
            & ROUGE-1  & - & - & - & - & - & - & - & - \\
            & ROUGE-L  & - & - & - & - & - & - & - & - \\\midrule
        \multirow{3}{*}{\textbf{NewsIT}}
            & BLEU     & 63.5 & 43.9 & 29.7& 45.8 & 29.8 & 0.2 & 43.9 & 64.4 \\
            & ROUGE-1  & - & -& - & - & - & - & - & - \\
            & ROUGE-L  & - & - & - & - & - & - & - & - \\\midrule
        \multirow{3}{*}{\textbf{StratQA}}
            & BLEU     & 60.7 & 5.0 & 4.6 & 7.5 & 6.4 & 7.9 & 4.2 & 63.3 \\
            & ROUGE-1  & 57.9 & 15.8 & 4.8 & 5.7 & 11.9 & 7.6 & 16.8 & 61.0 \\
            & ROUGE-L  & 54.6 & 15.4 & 3.2 & 3.6 & 7.9 & 5.4 & 16.3 & 56.9 \\\midrule
        \multirow{3}{*}{\textbf{TopChat}}
            & BLEU     & 32.0 & 26.9 & 18.4 & 30.1 & 28.5 & 0.0 & 37.6 & 29.0 \\
            & ROUGE-1  & 31.1 & 9.5 & 4.1 & 3.6 & 6.4 & 1.4 & 8.9 & 29.7 \\
            & ROUGE-L  & 28.3 & 8.9 & 3.1 & 2.0 & 3.9 & 0.4 & 8.5 & 26.9 \\\midrule
        \multirow{3}{*}{\textbf{ALPACA}}
            & BLEU     & 62.2 & 25.9 & 66.8 & 68.4 & 7.2 & 0.0 & 24.8 & 66.0 \\
            & ROUGE-1  & 57.2 & 22.7 & 16.4 & 17.9 & 17.0 & 11.3 & 27.9 & 63.9 \\
            & ROUGE-L  & 52.3 & 20.0 & 11.7 & 11.9 & 12.6 & 10.2 & 26.2 & 57.5 \\\midrule
            \multirow{3}{*}{\textbf{Avg.}}&BLEU&51.0&34.4&26.9&30.1&23.0&6.5&37.4&51.9\\
            &ROUGE-1&38.5&24.2&13.3&12.0&20.9&8.1&25.4&40.1\\
            &ROUGE-L&35.8&22.6&9.2&8.6&16.9&5.8&24&36.9\\
        \bottomrule
    \end{tabular}}
    \caption{The BLEU, ROUGE-1 and ROUGE-L results on 8 QA tasks by comparing different LoRA baselines under Qwen-2 1.5B as LLM backbone.}
    \label{table:generation_all_result_qwen_2_1.5b}
\end{table*}

\begin{table*}[tb]
    \centering
    \resizebox{\textwidth}{!}{
    \begin{tabular}{lcc|cccccc|c}
        \toprule
        \textbf{Models} & \textbf{Metric} & \textbf{PEFT (Ref.)} & \textbf{LoRA-F} & \textbf{TIES} & \textbf{DARE} & \textbf{Arrow} & \textbf{LoraHub} & \textbf{MeteoRA (T1-1k)} & \textbf{DLP-LoRA}\\
        \midrule
        
        \multirow{3}{*}{\textbf{CNNDM}}
            & BLEU     & 16.1 & 17.2 & 17.3 & 4.5 & 12.6 & 14.9 & 10.8 & 14.2 \\
            & ROUGE-1  & 16.9 & 25.2 & 14.6 & 12.0 & 14.6 & 8.0 & 22.0 & 15.5 \\
            & ROUGE-L  & 15.4 & 24.0 & 9.7 & 9.0 & 10.4 & 3.2 & 20.1 & 14.0 \\\midrule
        \multirow{3}{*}{\textbf{LingPuzz}}
            & BLEU     & 57.2 & 35.6 & 33.0 & 55.8 & 58.0 & 37.8 & 40.2 & 56.8 \\
            & ROUGE-1  & 47.8 & 65.4 & 40.2 & 33.8 & 70.0 & 22.5 & 67.2 & 46.7 \\
            & ROUGE-L  & 46.2 & 62.9 & 30.4 & 27.0 & 63.2 & 17.2 & 61.3 & 46.0 \\\midrule
        \multirow{3}{*}{\textbf{NewsDE}}
            & BLEU     & 63.6 & 75.8 & 44.9 & 34.2 & 35.6 & 10.3 & 83.6 & 68.8 \\
            & ROUGE-1  & - & - & - & - & - & - & - & - \\
            & ROUGE-L  & - & - & - & - & - & - & - & - \\\midrule
        \multirow{3}{*}{\textbf{NewsES}}
            & BLEU     & 68.9 & 78.5 & 27.9 & 15.4 & 30.0 & 0.1 & 79.0 & 66.9 \\
            & ROUGE-1  & -  & - & - & - & - & - & - & - \\
            & ROUGE-L  & - & - & - & - & - & - & - & - \\\midrule
        \multirow{3}{*}{\textbf{NewsIT}}
            & BLEU     & 69.6 & 52.8 & 36.9 & 51.0 & 36.8 & 0.4 & 52.6 & 65.1 \\
            & ROUGE-1  & - & - & - & - & - & - & - & - \\
            & ROUGE-L  & - & - & - & - & - & - & - & - \\\midrule
        \multirow{3}{*}{\textbf{StratQA}}
            & BLEU     & 67.8 & 9.0 & 9.1 & 12.7 & 11.0 & 12.7 & 8.9 & 68.0 \\
            & ROUGE-1  & 67.3 & 22.1 & 9.3 & 11.3 & 18.7 & 13.0 & 23.8 & 67.7 \\
            & ROUGE-L  & 65.0 & 20.1 & 8.0 & 8.7 & 14.8 & 10.1 & 22.1 & 65.6 \\\midrule
        \multirow{3}{*}{\textbf{TopChat}}
            & BLEU     & 33.6 & 32.0 & 24.6 & 36.9 & 33.1 & 0.2 & 43.9 & 34.8 \\
            & ROUGE-1  & 33.7 & 14.1 & 8.0 & 7.8 & 10.1 & 4.1 & 13.7 & 35.9 \\
            & ROUGE-L  & 31.7 & 13.0 & 6.7 & 5.8 & 8.0 & 2.7 & 13.0 & 33.9 \\\midrule
        \multirow{3}{*}{\textbf{ALPACA}}
            & BLEU     & 63.9 & 30.1 & 71.8 & 72.1 & 11.0 & 0.3 & 29.8 & 63.8 \\
            & ROUGE-1  & 61.5 & 27.0 & 20.1 & 21.6 & 21.0 & 16.2 & 33.5 & 61.2 \\
            & ROUGE-L  & 56.1 & 25.3 & 16.8 & 17.6 & 18.1 & 14.0 & 31.8 & 56.0 \\\midrule
            \multirow{3}{*}{\textbf{Avg.}}&BLEU&55.1&41.4&33.2&35.3&28.5&9.6&43.6&54.8\\
            &ROUGE-1&45.4&30.8&18.4&17.3&26.9&12.8&32.0&45.4\\
            &ROUGE-L&42.9&29.1&14.3&13.6&22.9&9.4&29.7&43.1\\
        \bottomrule
    \end{tabular}}
    \caption{The BLEU, ROUGE-1 and ROUGE-L results on 8 QA tasks by comparing different LoRA baselines under Qwen-2 7B as LLM backbone.}
    \label{table:generation_all_result_qwen_2_7b}
\end{table*}

\begin{table*}[tb]
    \centering
    \resizebox{\textwidth}{!}{
    \begin{tabular}{lcc|cccccc|c}
        \toprule
        \textbf{Models} & \textbf{Metric} & \textbf{PEFT (Ref.)} & \textbf{LoRA-F} & \textbf{TIES} & \textbf{DARE} & \textbf{Arrow} & \textbf{LoraHub} & \textbf{MeteoRA (T1-1k)} & \textbf{DLP-LoRA}\\
        \midrule
        
        \multirow{3}{*}{\textbf{CNNDM}}
            & BLEU     & 8.0 & 12.6 & 13.0 & 2.1 & 10.2 & 11.2 & 8.3 & 14.3 \\
            & ROUGE-1  & 7.4 & 19.3 & 10.2 & 8.9 & 10.4 & 4.7 & 15.2 & 13.2 \\
            & ROUGE-L  & 7.0 & 18.6 & 5.0 & 4.8 & 7.3 & 1.6 & 16.4 & 12.5 \\\midrule
        \multirow{3}{*}{\textbf{LingPuzz}}
            & BLEU     & 58.0 & 31.7 & 30.0 & 52.1 & 56.5 & 34.9 & 38.4 & 56.4 \\
            & ROUGE-1  & 45.4 & 60.1 & 37.6 & 30.0 & 65.3 & 20.1 & 63.1 & 43.9 \\
            & ROUGE-L  & 44.1 & 58.6 & 27.3 & 25.1 & 59.8 & 14.6 & 58.0 & 41.9 \\\midrule
        \multirow{3}{*}{\textbf{NewsDE}}
            & BLEU     & 69.4 & 70.1 & 40.3 & 31.2 & 31.5 & 8.3 & 79.3 & 67.6 \\
            & ROUGE-1  & - & - & - & - & - & - & - & - \\
            & ROUGE-L  & - & - & - & - & - & - & - & - \\\midrule
        \multirow{3}{*}{\textbf{NewsES}}
            & BLEU     & 68.7 & 72.7 & 24.2 & 11.7 & 26.8 & 0.0 & 75.7 & 67.0 \\
            & ROUGE-1  & - & - & - & - & - & - & - & - \\
            & ROUGE-L  & - & - & - & - & - & - & - & - \\\midrule
        \multirow{3}{*}{\textbf{NewsIT}}
            & BLEU     & 69.7 & 48.8 & 32.8 & 48.0 & 32.1 & 0.1 & 48.3 & 67.4 \\
            & ROUGE-1  & - & - & - & - & - & - & - & - \\
            & ROUGE-L  & - & - & - & - & - & - & - & - \\\midrule
        \multirow{3}{*}{\textbf{StratQA}}
            & BLEU     & 65.6 & 7.1 & 6.5 & 9.9 & 8.5 & 10.1 & 6.2 & 66.5 \\
            & ROUGE-1  & 59.9 & 18.5 & 6.7 & 8.4 & 14.7 & 10.0 & 19.7 & 60.1 \\
            & ROUGE-L  & 56.8 & 17.9 & 5.6 & 5.8 & 10.5 & 7.2 & 18.6 & 56.7 \\\midrule
        \multirow{3}{*}{\textbf{TopChat}}
            & BLEU     & 33.6 & 29.8 & 21.0 & 33.7 & 30.4 & 0.1 & 40.2 & 33.7 \\
            & ROUGE-1  & 32.2 & 12.4 & 6.3 & 5.8 & 8.4 & 3.0 & 11.3 & 30.2 \\
            & ROUGE-L  & 30.2 & 11.5 & 5.1 & 4.8 & 6.4 & 1.8 & 11.4 & 28.3 \\\midrule
        \multirow{3}{*}{\textbf{ALPACA}}
            & BLEU     & 64.7 & 28.4 & 69.2 & 70.0 & 9.3 & 0.0 & 27.6 & 66.4 \\
            & ROUGE-1  & 59.2 & 25.6 & 18.3 & 19.5 & 19.0 & 14.3 & 30.8 & 61.7 \\
            & ROUGE-L  & 53.6 & 22.9 & 14.2 & 14.8 & 15.6 & 12.0& 28.8 & 55.9 \\\midrule
            \multirow{3}{*}{\textbf{Avg.}}&BLEU&54.7&37.7&29.6&32.3&25.7&8.1&40.5&54.9\\
            &ROUGE-1&40.8&27.2&15.8&14.5&23.6&10.4&28.0&41.8\\
            &ROUGE-L&38.3&25.9&11.4&11.1&19.9&7.4&26.6&39.1\\
        \bottomrule
    \end{tabular}}
    \caption{The BLEU, ROUGE-1 and ROUGE-L results on 8 QA tasks by comparing different LoRA baselines under LLaMA-2 7B as LLM backbone.}
    \label{table:generation_all_result_llama_2_7b}
\end{table*}

\begin{table*}[tb]
    \centering
    \resizebox{\textwidth}{!}{
    \begin{tabular}{lcc|cccccc|c}
        \toprule
        \textbf{Models} & \textbf{Metric} & \textbf{PEFT (Ref.)}  & \textbf{LoRA-F} & \textbf{TIES} & \textbf{DARE} & \textbf{Arrow} & \textbf{LoraHub} & \textbf{MeteoRA (T1-1k)} & \textbf{DLP-LoRA}\\
        \midrule
        
        \multirow{3}{*}{\textbf{CNNDM}}
            & BLEU     & 9.0 & 16.1 & 18.1 & 4.7 & 13.1 & 15.3 & 11.9 & 17.9 \\
            & ROUGE-1  & 9.7 & 24.8 & 15.4 & 13.7 & 15.3 & 8.7 & 23.3 & 18.9 \\
            & ROUGE-L  & 8.8 & 23.3 & 10.9 & 9.6 & 11.1 & 3.8 & 21.8 & 17.8 \\\midrule
        \multirow{3}{*}{\textbf{LingPuzz}}
            & BLEU     & 64.6 & 36.9 & 34.2 & 56.2 & 59.0 & 39.3 & 41.7 & 65.7 \\
            & ROUGE-1  & 58.6 & 71.8 & 43.2 & 35.7 & 72.1 & 24.5 & 69.5 & 58.9 \\
            & ROUGE-L  & 57.2 & 66.6 & 33.9 & 28.1 & 65.9 & 18.4 & 63.6 & 58.1 \\\midrule
        \multirow{3}{*}{\textbf{NewsDE}}
            & BLEU     & 68.2 & 78.3 & 46.5 & 36.6 & 37.4 & 11.9 & 86.5 & 58.7 \\
            & ROUGE-1  & -  & - & - & - & - & - & - & - \\
            & ROUGE-L  & -  & - & - & - & - & - & - & - \\\midrule
        \multirow{3}{*}{\textbf{NewsES}}
            & BLEU     & 69.1 & 81.5 & 30.6 & 17.6 & 31.8 & 0.0 & 81.5 & 69.2 \\
            & ROUGE-1  & -  & - & - & - & - & - & - & - \\
            & ROUGE-L  & -  & - & - & - & - & - & - & - \\\midrule
        \multirow{3}{*}{\textbf{NewsIT}}
            & BLEU     & 65.6 & 54.9 & 37.5 & 52.2 & 38.0 & 0.0 & 54.9 & 68.4 \\
            & ROUGE-1  & -  & - & - & - & - & - & - & - \\
            & ROUGE-L  & -  & - & - & - & - & - & - & - \\\midrule
        \multirow{3}{*}{\textbf{StratQA}}
            & BLEU     & 64.3 & 10.4 & 10.9 & 14.8 & 12.2 & 14.4 & 10.6 & 66.2 \\
            & ROUGE-1  & 62.8 & 23.4 & 10.2 & 12.8 & 20.6 & 14.7 & 25.2 & 63.5 \\
            & ROUGE-L  & 60.0 & 22.3 & 8.2 & 10.0 & 16.5 & 11.6 & 23.9 & 60.1 \\\midrule
        \multirow{3}{*}{\textbf{TopChat}}
            & BLEU     & 36.0 & 33.8 & 26.1 & 38.3 & 35.6 & 0.1 & 45.6 & 29.6 \\
            & ROUGE-1  & 35.9 & 15.0 & 9.2 & 8.6 & 11.2 & 4.9 & 15.2 & 30.3 \\
            & ROUGE-L  & 33.5 & 14.0 & 7.7 & 6.6 & 9.1 & 3.1 & 14.1 & 27.8 \\\midrule
        \multirow{3}{*}{\textbf{ALPACA}}
            & BLEU     & 64.4 & 31.5 & 73.5 & 73.5 & 12.3 & 0.0 & 32.3 & 63.4 \\
            & ROUGE-1  & 61.8 & 28.4 & 21.4 & 23.0 & 22.2 & 17.6 & 35.8 & 61.2 \\
            & ROUGE-L  & 56.6 & 26.7 & 18.1 & 19.2 & 18.6 & 15.1 & 33.5 & 56.3 \\\midrule
            \multirow{3}{*}{\textbf{Avg.}}&BLEU&55.2&42.9&34.7&36.7&29.9&10.1&45.6&54.9\\
            &ROUGE-1&45.8&32.7&19.9&18.8&28.3&14.1&33.8&46.6\\
            &ROUGE-L&43.2&30.6&15.8&14.7&24.2&10.4&31.4&44.0\\
        \bottomrule
    \end{tabular}}
    \caption{The BLEU, ROUGE-1 and ROUGE-L results on 8 QA tasks by comparing different LoRA baselines under LLaMA-3 8B as LLM backbone.}
    \label{table:generation_all_result_llama_3_8b}
\end{table*}

\begin{figure*}[tb]
    \centering
    \includegraphics[width=\textwidth]{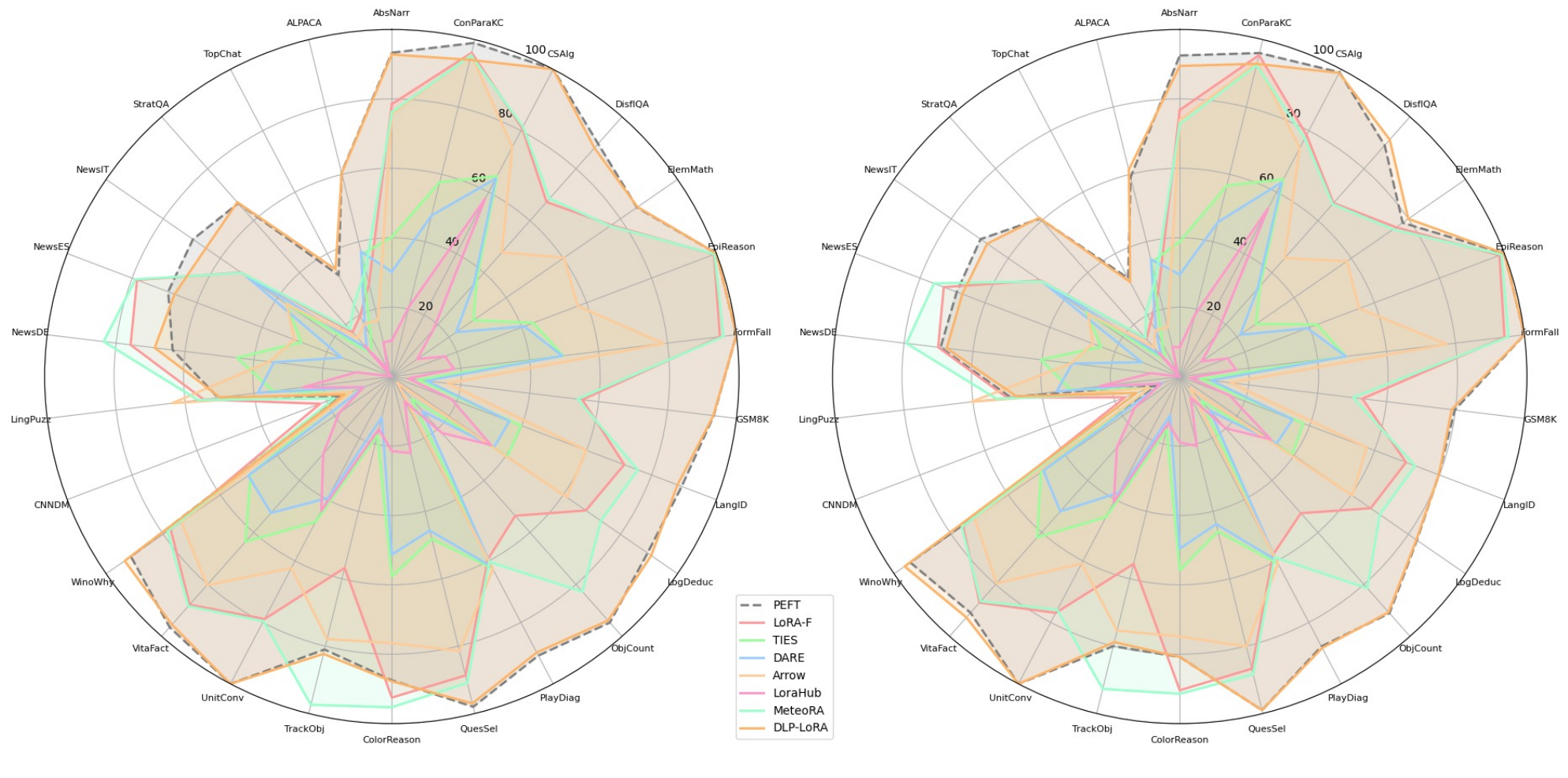}
    \caption{Radar chart of Qwen-2 7B and LLaMA-2 7B across 18 MCQ and 8 QA tasks.}
    \label{fig:radar_qwen2_7b_llama_3_7b}
\end{figure*}

We further demonstrate more radar charts to show more results for Qwen-2 7B and LLaMA-2 7B backbones in Figure~\ref{fig:radar_qwen2_7b_llama_3_7b}. 

\section{Composite-$n$ Task Results across Four LLMs Backbones}\label{app:composite}
Table~\ref{tab:app_composite_n} shows all details about composite-$n$ tasks by comparing the Basic LLMs, LoRA-F ($r=64$) and our DLP-LoRA under composite-$18$ and composite-$8$ task settings.
\begin{table*}[tb]
    \centering
    \small
    \resizebox{\textwidth}{!}{
    \begin{tabular}{llcccc}
    \toprule
    \textbf{Model} & \textbf{Method} &  \textbf{Acc.} (\%) $\uparrow$& \textbf{BLEU} $\uparrow$& \textbf{ROUGE-1} $\uparrow$& \textbf{ROUGE-L} $\uparrow$\\\midrule
    \multirow{3}{*}{\textbf{Qwen-2 1.5B}} & Basic
          &31.65&51.48&48.69&45.72\\
         & LoRA-F ($r=64$)  & 33.23 & 51.46 & 48.86 & 45.90\\
         &DLP-LoRA &\textbf{90.43}&\textbf{56.00}&\textbf{54.61}&\textbf{52.27}\\
         \midrule
    \multirow{3}{*}{\textbf{Qwen-2 7B}} & Basic
          &58.59&53.25&50.70&48.58\\
         & LoRA-F ($r=64$)  & 59.42 & 53.63 & 51.75 & 48.92\\
         &DLP-LoRA &\textbf{92.75}&\textbf{57.44}&\textbf{56.84}&\textbf{54.90}\\
         \midrule
    \multirow{3}{*}{\textbf{LLaMA-2 7B}} & Basic
          &36.29&52.32&46.78&44.36\\
         & LoRA-F ($r=64$)  & 37.93 & 52.84 & 46.96 & 45.35\\
         &DLP-LoRA &\textbf{91.20}&\textbf{58.61}&\textbf{54.70}&\textbf{52.60}\\
         \midrule
    \multirow{3}{*}{\textbf{LLaMA-3 8B}} & Basic
          &65.44&52.00&50.16&47.16\\
         & LoRA-F ($r=64$)  & 65.98 & 52.26 & 50.38 & 47.40\\
         &DLP-LoRA &\textbf{96.03}&\textbf{57.79}&\textbf{57.45}&\textbf{55.35}\\\midrule
    \multirow{3}{*}{\textbf{Avg.}} & Basic & 47.99 & 52.26& 49.08& 46.46\\
         & LoRA ($r=64$)  & 49.14 & 52.55 & 49.49 & 46.89\\
    & DLP-LoRA  & $\bm{92.60_{\textcolor{teal}{+92.95\%}}}$ & $\bm{57.46_{\textcolor{teal}{+9.95\%}}}$& $\bm{55.90_{\textcolor{teal}{+13.90\%}}}$& $\bm{53.78_{\textcolor{teal}{+15.76\%}}}$\\
    \bottomrule
    \end{tabular}}
    \caption{Evaluation results for composite-$n$ task, where composite-8 includes all QA tasks, and composite-18 includes all MCQ tasks. In addition, we compare a single LoRA with a higher rank trained on composite-26 task setting. The evaluation results are averaged after running 10 times. The subscript percentage denotes relative accuracy, BLEU, ROUGE-1 and ROUGE-L improvement or reduction over each basic LLMs baseline.}
    \label{tab:app_composite_n}
\end{table*}